\def\year{2022}\relax
\documentclass[letterpaper]{article} 
\usepackage{aaai22}  
\usepackage{times}  
\usepackage{helvet}  
\usepackage{courier}  
\usepackage[hyphens]{url}  
\usepackage{graphicx} 
\usepackage{natbib}  
\usepackage{caption} 
\DeclareCaptionStyle{ruled}{labelfont=normalfont,labelsep=colon,strut=off} 
\usepackage{algorithm}
\usepackage{algorithmic}

\usepackage{newfloat}
\usepackage{listings}
\floatstyle{ruled}
\newfloat{listing}{tb}{lst}{}
\floatname{listing}{Listing}
\title{Scalable Rail Planning and Replanning with Soft Deadlines}
\author{Zhe Chen\textsuperscript{\rm 1},
        Jiaoyang Li\textsuperscript{\rm 2},
        Daniel Harabor\textsuperscript{\rm 1},
        Peter J. Stuckey\textsuperscript{\rm 1}
        \\}
\affiliations{\textsuperscript{\rm 1}Monash University, Australia \\
             \textsuperscript{\rm 2}University of Southern California, USA \\
             \{zhe.chen,daniel.harabor,peter.stuckey\}@monash.edu, jiaoyanl@usc.edu}

\usepackage{color}

\begin{document}

\maketitle

\begin{abstract}
The Flatland Challenge, which was first held in 2019 and reported in NeurIPS 2020, is designed to answer the question: How to efficiently manage dense traffic on complex rail networks?
Considering the significance of punctuality in real-world railway network operation and the fact that fast passenger trains share the network with slow freight trains, Flatland version 3 introduces trains with different speeds and scheduling time windows.
This paper introduces the Flatland 3 problem definitions 
and extends an award-winning MAPF-based
software, which won the NeurIPS 2020 competition, to efficiently solve Flatland 3 problems. The resulting system won the Flatland 3 competition. 
We designed a new priority ordering for initial planning, a new neighbourhood selection strategy for efficient solution quality improvement with Multi-Agent Path Finding via Large Neighborhood Search(MAPF-LNS), and use MAPF-LNS for partially replanning the trains influenced by malfunction.

\end{abstract}

\section{Introduction}
The Flatland 3 Challenge is the third edition of this popular railway network operation competition.
The competition is organized by AICrowd, SBB (Swiss federal railways), SNCF (French national railway company), and Deutsche Bahn (German national railway company), which aims to answer the question:``How to efficiently manage dense traffic on complex rail networks?''.

The challenge was first held in 2019, where most participants used planning or operation research methods to solve the problem. 
To encourage participants to use reinforcement learning approaches, the NeurIPS 2020 Flatland Challenge \cite{laurent2021flatland} 
added a separate reinforcement learning track (which ranks only reinforcement learning approaches). 
The competition format changed to tackle an unbounded number of instances of increasing difficulty in 8 hours, in anticipation that the 
computation speed and large problem size would be bottlenecks for non-reinforcement learning approaches.
However, the 2020 competition shows no doubt that planning-based approaches again dominated reinforcement learning approaches.

The challenge simulates railway network operations on an idealized railway network, a grid-based map showing rail tracks and train stations with a set of trains with start and target stations. Our task is to navigate trains to their target stations while complying with the rules of the rail transaction and avoiding collisions.
The Flatland 3 Challenge introduces more elements from real-world rail operations. One change reflects the fact that timing and punctuality are crucial for real-world railways. The challenge schedules a time window for each train: the earliest departure time and an expected arrival time. 
The other change reflects how fast passenger trains and slow freight trains share the same railway network in the real world.
The challenge considers trains with different speed profiles, modelled by the minimal number of time steps needed to travel through a rail segment.

A significant hidden challenge of the competition 
is to overcome the slow execution speed of the competition environment.
In the NeurIPS 2020 Flatland challenge, the winning software only used $30\%$ of the $8$ hours for planning agents, the rest was spent in executing the simulation environment. 
For Flatland 3, the total planning time is only $5\%$ of the total time, meaning 
over 2 hours of evaluation time, we only have about 7 minutes for planning. 
In the challenge, reducing the makespan (the time the last train arrived) also helps reduce the environment execution time, since it executes fewer steps, but in Flatland 3, because of the earliest departure times, the overall makespan is hard to reduce by planning (it is strongly bounded by the latest leaving trains).  These considerations mean that a careful balance is required in how much time should be spent improving the plans. Any time-consuming optimisation leads to a reduction in the number of solved problems, which can cause significant score loss but limited per-instance score improvement. 

The challenge is highly related to the academic problem of Multi-Agent Path Finding(MAPF). MAPF defines a graph and a group of agents, where each agent has a start and target vertex, and we need to plan collision-free paths for all agents while minimizing an objective, e.g. sum of individual costs. 
The problem is essential for a wide range of applications, including computer games~\cite{SigurdsonCIG18, LiAAMAS20a}, automated warehousing~\cite{MaAAMAS17lifelong,chen2021integrated,LiAAMAS20b}, UAV traffic management~\cite{HoAAMAS19} and drone swarms~\cite{HoenigTRO18}. 
Variants of the MAPF problem, such as MAPF with motion planning \cite{CohenSoCS19}, MAPF with deadlines \cite{ma2018deadline} and MAPF with delay probabilities\cite{CapIROS16,chen2021symmetry,MaAAAI17delay,LiAIAA19,WagnerICAPS17,AtzmonICAPS20} are also widely studied and closely related to the Flatland environment.

In this paper, we introduce the definition of the Flatland 3 problem, illustrate how Flatland 3 differs from previous editions, and describe a MAPF-based software that efficiently plans and replans punctual paths for trains with different speeds, winning the Flatland 3 competition.

\section{Flatland 3 Environment}

\subsection{Problem Definition}
\begin{figure}
    \centering
    \includegraphics[width=0.8\columnwidth]{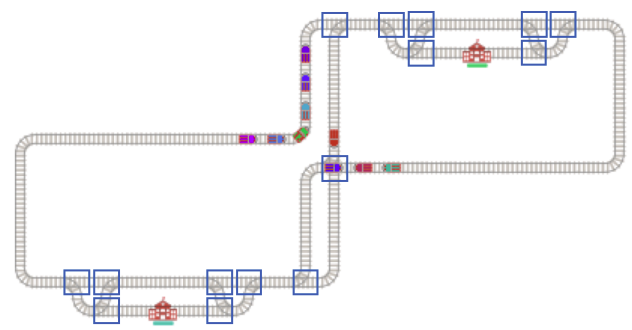}
    \caption{Example of flatland railway network. \cite{laurent2021flatland}}
    \label{fig:rail}
\end{figure}

\begin{figure}[t]
\centering
\begin{tabular}{cccccccc}
\includegraphics[width=0.06\columnwidth]{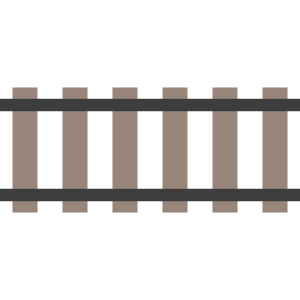} &
\includegraphics[width=0.06\columnwidth]{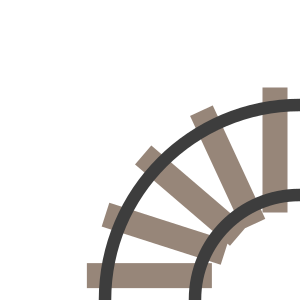} &
\includegraphics[width=0.06\columnwidth]{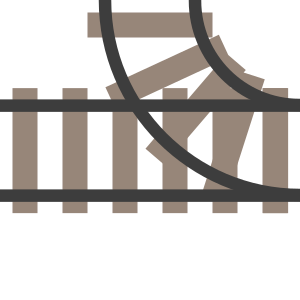} &
\includegraphics[width=0.06\columnwidth]{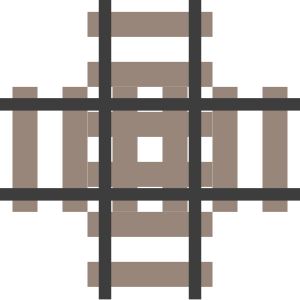} &
\includegraphics[width=0.06\columnwidth]{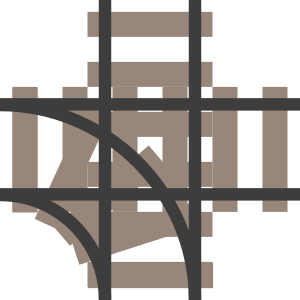} &
\includegraphics[width=0.06\columnwidth]{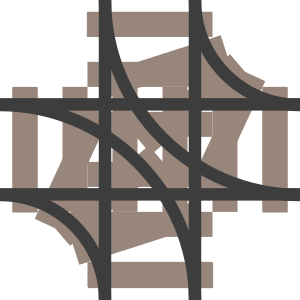} &
\includegraphics[width=0.06\columnwidth]{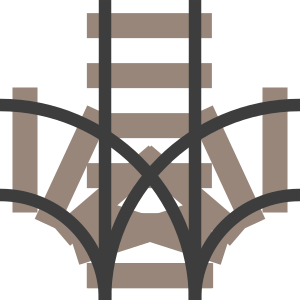} &
\includegraphics[width=0.06\columnwidth]{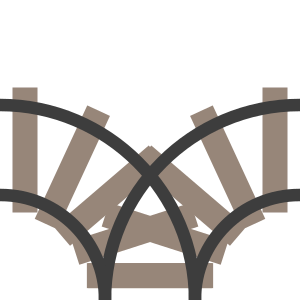} \\
(a) & (b) & (c) & (d) & (e) & (f) & (g) & (h) \\
\end{tabular}
\caption{Flatland rail types: (a) straight, (b) curve, (c) simple switch, (d) diamond crossing, (e) single slip switch, (f) double slip switch, (g) tri-symmetrical switch, and (h) symmetrical switch. \cite{li2021scalable}}
\label{fig:railcell}
\end{figure}

Flatland 3 aims to solve the rail planning and replanning problem with soft deadlines based on a railway network represented by a $w \times h$ grid map with $n$ cities, where each traversable cell is associated with a rail type shown in Figure \ref{fig:railcell}. 
The rail type determines how trains can traverse through the cell.
Each city is a small area on the map and has an even number (minimum 2) of parallel rails, where one rail in the city contains a train arrival station, and the other rail contains a departure station. Figure \ref{fig:rail} shows an example with 2 cities, where the red building indicates an arrival station and the departure station is the adjacent cell on the other rail.
Time is discretized into timesteps from 0 to $T_{max}=\lfloor8(w + h + \frac{m}{n})\rfloor$.
There is a set of $m$ trains $\{a_1, a_2, ..., a_m\}$ in a problem. 
Each agent $a_i$ has a start cell $s_i$ (= a train station), 
an initial orientation $d_i$, a target cell $g_i$ (= another train station), a max speed counter $ C^{max}_i \in [1 ... 4]$ (indicate the minimum timesteps needed to traverse through a cell), the earliest departure timestep $EDT_i$, and an expected arrival timestep $EAT_i$ (a soft deadline). Note that the speed counter $C^{max}$ is an inverse of speed: $C^{max} = 1$ trains can move in every time step, $C^{max} = 4$
trains can move at most once in 4 timesteps.

Our task is to navigate as many trains as possible to their target cells and minimize total arrival delays for those who did not catch EAT.
To be more specific, we want to maximize the \emph{normalized reward} (or \emph{reward} for short) defined as:
    $1-\frac{\sum_{1 \leq i \leq m}D_i}{mT_{max}} \in [0, 1]$, 
where $D_i = min(EAT_i - ACT_i,0)$ is how many timesteps $a_i$ is delayed arriving at its goal $g_i$, $ACT_i$ is the actual arrival time of agent $a_i$. If $a_i$ does not arrive $g_i$ before $T_{max}$, $ACT_i$ is estimated as $T_{max} + distance(v_i, g_i)$ ($v_i$ is the location of $a_i$ at $T_{max}$ or $s_i$ if $a_i$ does not enter the environment at $T_{max}$).
We define \emph{success rate} as the percentage of trains that reach their target cells by timestep $T_{max}$ and \emph{earliest arrival time} $T_i^0$ of the train $a_i$ as the earliest timestep when it can reach its target cell when ignoring collisions with other agents.

Each agent is parked off map at timestep 0 and leaves the map immediately when they arrive at its goal cell. 
An agent $a_i$ appears in its start cell (enter the map) with its initial orientation and a speed counter $C_i = 0$ when receiving a move forward command in or after timestep $EDT_i$.
At each timestep, an agent only occupies one cell, and we navigate all agents giving each of them a command.
When $a_i$ is on the map, a move forward/left/right command will:
\begin{itemize}
    \item increase $C_i$ by $1$ if $C_i < C^{max}_i$,
    \item set $C_i = 0$ and move $a_i$ to an adjacent cell following the transition rule of a rail type if $C_i == C^{max}_i$, and the move action does not collide with any other trains and $a_i$ is not suffering from a malfunction,
    \item keep $C_i$ unchanged and stay in the current cell otherwise.
\end{itemize}
Two actions collide iff two agents arrive at the same cell or two agents swap adjacent cell locations at the same timestep. A stop command leaves the agent's location and $C$ unchanged. 
Malfunctions simulate delays by stopping a train at a random timestep for a random duration. The random timestep is generated by a Poisson process with a rate $\lambda$. The random duration is uniformly selected from the positive integer range and the delay duration becomes known when the malfunction occurs. The value of $\lambda$ and the range of random delay duration is known to the planning code.

The competition provides a software library written in Python to simulate the environment. For each instance, the simulation ends when all agents reach their target cells or reach $T_{max}$. Our solution is written as a C++ dynamic library and called by the Python simulator. As discussed in the introduction, this environment consumes the majority of the execution time. 

\subsection{Competition Configuration}
The challenge evaluates participants' codes in 150 instances with a time limit of 2 hours. These instances are categorized into 15 difficulty levels and each level contains 10 distinct instances. The easiest level has $30 \times 30$ grid maps with 7 agents and 2 cities. The hardest level has $158 \times 158$ grid maps with 425 agents and 41 cities.

\subsection{Flatland Challenge and MAPF}

\cite{li2021scalable} showed the Flatland Challenge has important differences from standard MAPF, but is closely related to MAPF variants. 
In the standard MAPF definition, we navigate a team of agents from their start vertices to goal vertices on an undirected graph with minimum flow time and without collisions. At each timestep, agents can move to an adjacent vertex or wait at the current vertex. 
Unlike standard MAPF, the Flatland environment restricts train movement to rails, trains are parked off the map before entering the map, and after reaching their target, the maximum time $T_{max}$ acts as a hard deadline for all trains and trains break down randomly while moving.

Flatland 3 further restricts trains' movement based on a speed profile, adds soft deadlines and departure times for each train, 
and uses the objective of minimizing arrival delays.

\section{NeurIPS 2020 Flatland Challenge}

Compared to the NeurIPS 2020 Flatland Challenge, the organizer of the competition introduced more elements from real-world railway operations.
In the NeurIPS 2020 Flatland Challenge, all trains have the same speed $C^{max} = 1$, have no earliest departure timestep or expected arrival timestep, and the optimisation objective is similar to the sum of individual costs, e.g., the total number of steps taken by all trains to reach their targets. 
In addition, the 2020 challenge evaluates solutions over an infinite number of instances with increasing difficulty with an 8 hour runtime limit. 

Just as in Flatland 3, the 2020 challenge has an overall track, which takes all solutions into account, and a reinforcement learning track, which only considers reinforcement learning-based solutions. 
The winning solution of the overall track, a MAPF-based approach, solved 362 instances with a score of 297.507 and the largest instance contains 3,256 trains. The top solution from the reinforcement learning track spent 8 hours solving 336 instances with a score of 214.150, which is reached by the top MAPF-based solution after only 15 minutes.

\subsection{MAPF-based Flatland 2020 solution}

\cite{li2021scalable} developed a MAPF-based software, which incorporates many state-of-the-art MAPF techniques, for solving train planning and replanning problems on large-scale networks under uncertainty and 
won the NeurIPS 2020 Flatland Challenge. 
Our solution for Flatland 3 is based on this work.

Their basic solution uses \emph{Prioritized Planning}~(PP)~\cite{WHCA} to generate the initial solution and uses \emph{Minimal Communication Policy}~(MCP)~\cite{MaAAAI17delay} to handle malfunctions during execution. 
\emph{MAPF via Large Neighborhood Search}~(MAPF-LNS)~\cite{li2021anytime}, \emph{Partial Replanning}~(PR) and \emph{Lazy Planning}~(LP) \cite{li2021scalable} further improved their solution.

PP first sorts agents in a priority order, from high priority to low priority.
Then uses \emph{Safe Interval Path Finding}(SIPP)\cite{SIPP} to plan the shortest paths while avoiding collisions with already planned paths, for each agent in the priority order. 
The Flatland Environment problems are well-formed\cite{MaAAAI19}, hence PP is guaranteed to find a solution if such a solution exists. Although the PP computes solutions rapidly, its solution quality is far from optimal.

MAPF-LNS further improves the solution of PP by repeating a Large Neighborhood Search process to improve the quality. 
It takes the solution of PP as input and repeatedly selects, destroys, and replans the paths of a subset of agents until an iteration limit is reached. 
Li~\emph{et al} runs 4 PP with different priority orders followed 
by 4 MAPF-LNS processes in parallel to select the best solution. 
To balance the trade-off between solution quality and runtime they collected training data offline and use \emph{Simulated Annealing} to determine the LNS iteration limits for instances of different sizes.

MCP stops some trains to maintain the order in which each train 
visits shared cells to avoid potential deadlocks caused by malfunctions.
But it sometimes stops trains unnecessarily. 
They designed a PR mechanism, which selects and replans the paths of agents that are influenced by malfunctioning agents, to overcome this issue. 

When there are thousands of agents to schedule, the runtime of PP with SIPP grows rapidly as it has to plan paths that avoid collisions with an increasing number of existing paths. The LP scheme tackles this issue by only planning paths for some of the agents during the initial planning phase and planning the rest during the execution. 
It prevents pushing too many agents into the environment, thus avoiding severe traffic congestion, taking into account malfunctions that have already happened in later planning, ignoring the paths of finished agents and significantly reducing the planning runtime.

\section{Rail Planning and Replanning with Soft Deadlines}

In this section, we introduce a modified and improved version of \cite{li2021scalable}'s solution which solves Flatland 3 problems efficiently. 
We evaluated our solution over 150 locally generated instances, in which we simulate the challenge benchmark based on public challenge configuration, on a Nectar Cloud Server with AMD Opteron 63xx CPU and 32 GB RAM.
The source codes and evaluation instances will be made public upon publication.

\subsection{SIPP with Discrete Speed}

One advantage of SIPP is that it is capable of planning paths with motion constraints~\cite{ma2019lifelongkinematic}. A search node $n = \langle v, I, t \rangle$ in SIPP includes a current vertex $v$, an obstacle-free time interval $I=[l, u)$ of $v$, and an earliest possible arrival time $t$. 
SIPP expands $n$ by generating successor nodes for all reachable vertexes' reachable time intervals from $n$, where $n'$ is a successor of $n$,  $I'$ is the interval of $n'$, $I'.l <= I.u$, $I'.u > n.t + 1$ and $n'.t = max(n.t+1, I'.l)$.

The nature of discrete speed in Flatland 3 is a kind of motion constraint that, an agent $a_i$ must stay at a vertex for at least $C^{max}_i$ timestep before it traverses to the next vertex. To satisfy this constraint, we generate a successor node $n'$ iff $I'.u > n.t + C^{max}_i$. Then the earliest possible arrival time of $n't$ is $max(n.t + C^{max}_i, I'.l)$.  

Our basic Flatland 3 solution uses SIPP with discrete speed for PP, modifies the SIPP to only allow agents entering the map at or after $EDT$ of each agent, disables LP as there are at most 425 agents in any competition instance, disables PR to solve more problems within the time limit, and optimized the coding quality. 
The LNS is modified to accept a replanned solution iff the total arrival delay is improved in each iteration, and the iteration limit is at most 50 for small and large instances and at most 500 for instances in the middle, as we observed locally that the instances in the middle perform worse than others without LNS. 
This solution solves 135 problems in 2 hours and gives a score of 123.966. But this is not enough to win the competition. 
Due to the increasing difficulty of evaluation instances and the long environment execution time on large instances, solving one more instance or improving the average score over all solved instances becomes extremely difficult. Hence the improvements we discuss below, why they seem tiny are in fact significant.

\subsection{Slack Based Priority}

In the 2020 solution, PP sorts agents by train index, earliest arrival time, or start cells. Using parallel computing, PP computes with different priority orders at the same time and selects the best solution. 
In the Flatland 3 challenge, the introducing of the soft deadline and the new optimization objective make the soft deadline an important factor for priority ordering.

A basic idea would be to prioritize agents with the earliest soft deadline, 
but this can be misleading. Assuming one agent $a_1$ has a late $EAT_1$, but also a late  $EDT_1$. The shortest path distance between the start and goal vertex might be equal to $EAT_1 - EDT_1$. 
Another agent $a_2$, which must collide with $a_1$, has $EAT_2 < EAT_1$, but the $EAT_2 - EDT_2$ is far larger than the length of its shortest path. 
Clearly, giving $a_1$ higher priority is a better choice, although it has a later $EAT_1$.
Considering the scenario above, we define $slack_i = EAT_i - EDT_i - distance(s_i,g_i)$ of agent $a_i$ as a better metric for a priority order. 

We define a new priority ordering based on $slack$ tie-breaking by prioritizing fast agents over slow agents. 
We added this priority order in the parallel-PP approach of the 2020 solution and it solved 135 problems with a score of 124.227. 

\subsection{Delay-based neighbourhood selection}

Neighbourhood selection, where the LNS select a subset of agents for replanning, is the key for MAPF-LNS to improve solutions efficiently. 
The 2020 solution designed three neighbourhood selection strategies: (1) an agent-based strategy, which selects a train that is heavily delayed and other trains that cause the delay, 
(2) an intersection-based strategy, which selects trains that visit the same intersection, (3) a start-based strategy, which selects trains with the same start cell. It uses adaptive LNS~\cite{ALNS} to keep track of each strategy's relative improvement and choose strategies randomly in proportion to their improvement.

Here we propose a delay-based strategy that takes account of soft deadlines. 
We randomly select an agent from all agents that can not arrive at its goal vertex before its expected arrival time and find other agents potentially blocking its way. 
Then we randomly prioritize the selected agents, replan their paths, and only accept their new path if this results in lower total arrival delays.

If we replace the start-based strategy (which is not relevant to Flatland 3 where trains have a scheduled departure time) in the adaptive LNS this improves the score to 124.352. If we disable adaptive LNS and use only delay-based neighbourhood selection we get a score of 124.432. We assume the relatively small iteration limit settings make it harder for adaptive LNS to learn the right balance of strategies.

\subsection{Partial Replanning using LNS}

PR fixes unnecessary waits caused by MCP when a new malfunction happens, but it also has the following drawbacks: (1) malfunction happens almost on every timestep on a large instance, and running PR on every timestep slows down the execution process, (2) too many trains need to be replanned on large instances, and (3) it replans paths without the guidance of the reward, in other words, if we can only replan a small proportion of affected agents, it does not know who to plan first.

To overcome these issues, we use MAPF-LNS for PR, which focuses on agents causing arrival delays and optimises total arrival delays during the execution. 
LNS PR also overcomes the issue that malfunctions during execution make the optimisation of the initial plan less effective.
Different from PR in the 2020 solution that is triggered by each new malfunction, we run LNS PR for a fixed number of times $r$ for each instance and each run has $p$ LNS iterations.  In other words, we run LNS PR every $\frac{T_{max}}{r}$ timesteps, 
rather than every timestep where a malfunction occurs. $r$ and $p$ are two integer parameters we configure for submissions. 
In this manner, we can balance the trade-off between solution quality and problem-solving speed better, and the influence from all malfunctions happening before each run is addressed.
By setting $r=20$ and $p=20$ we increase the score to 125.175. In comparison, if we trigger the standard PR with the same frequency, we get a score of 124.911.

\section{Conclusion}

The Flatland 3 challenge provides a chance to tackle a (simplified form of a) 
real-world problem.
The challenge result shows that MAPF-based approaches remain far ahead of Reinforcement Learning for this problem. 
Our software efficiently plans and optimises paths for hundreds of agents in seconds, while satisfying the speed and time window constraints, and delivers high-quality plan execution under uncertainty. 
On the official leader board, our winning solution reached a score of $135.47$ and solved $145$ instances. 
The second team on the leaderboard reached a score of $132.470$, we reach the same score in $1$ hour $50$ minutes (illustrating how a small difference in score actually represents a large change). The best reinforcement learning approach ended with a score of $27.868$ which we reach in only $3$ minutes.
Interestingly our score jumped to $140.99$ solving all $150$ instances after the organizers re-ran our solution on a faster computer after the competition finished.
We then beat the second team's score (on the same computer) in $1$ hour and $32$ minutes. 

\bibliography{main}

\end{document}